\documentclass[10pt,twocolumn,letterpaper]{article}

\usepackage[pagenumbers]{cvpr} 
\usepackage{graphicx}
\usepackage{amsmath}
\usepackage{amssymb}
\usepackage{booktabs}

\usepackage{siunitx}
\usepackage{multirow}
\usepackage[pagebackref,breaklinks,colorlinks]{hyperref}
\usepackage{float}

\usepackage[capitalize]{cleveref}
\crefname{section}{Sec.}{Secs.}
\Crefname{section}{Section}{Sections}
\Crefname{table}{Table}{Tables}
\crefname{table}{Tab.}{Tabs.}

\def\cvprPaperID{5524} 
\def\confName{CVPR}
\def\confYear{2023}

\newcolumntype{M}[1]{>{\centering\arraybackslash}m{#1}}

\begin{document}

\title{Generic Event Boundary Detection in Video with Pyramid Features}

\author{Van Thong Huynh, Hyung-Jeong Yang, Guee-Sang Lee, Soo-Hyung Kim\thanks{Corresponding author}\\
Department of AI Convergence, Chonnam National University\\
{\tt\small \{vthuynh,hjyang,gslee,shkim\}@jnu.ac.kr}
}
\maketitle

\begin{abstract}
   Generic event boundary detection (GEBD) aims to split video into chunks at a broad and diverse set of actions as humans naturally perceive event boundaries. In this study, we present an approach that considers the correlation between neighbor frames with pyramid feature maps in both spatial and temporal dimensions to construct a framework for localizing generic events in video. The features at multiple spatial dimensions of a pre-trained ResNet-50 are exploited with different views in the temporal dimension to form a temporal pyramid feature map. Based on that, the similarity between neighbor frames is calculated and projected to build a temporal pyramid similarity feature vector. A decoder with 1D convolution operations is used to decode these similarities to a new representation that incorporates their temporal relationship for later boundary score estimation. Extensive experiments conducted on the GEBD benchmark dataset show the effectiveness of our system and its variations, in which we outperformed the state-of-the-art approaches. Additional experiments on TAPOS dataset, which contains long-form videos with Olympic sport actions, demonstrated the effectiveness of our study compared to others.
\end{abstract}

\section{Introduction}

Temporal action localization attempts to spot the range of action instances in timestamps from the video stream with well-known including THUMOS~\cite{THUMOS14}, ActivityNet~\cite{Heilbron2015}, HACS~\cite{Zhao2019}, FineAction~\cite{Liu2021}. However, with the growth of video contents, the number of classes is expanding, and the predefined target classes could not cover completely. That can be resolved with semi-supervised or unsupervised learning, but it still focuses on specific actions. To increase generalization, Shou~\etal~\cite{Shou2021} constructed a new benchmark, Generic Event Boundary Detection (GEBD), which increases the ability to detect a board and diverse set of boundaries between taxonomy-free events, as humans naturally perceive.
Following cognitive studies~\cite{barker1955midwest}, GEBD~\cite{Shou2021} considered four high-level causes that involve changes in subject, action, shot, environment, and object of interaction to form event boundaries. These changes can occur with varying speed, duration and even do not involve any scene changes.

The area of GEBD is attracting considerable interest because of its application in the field of video understanding. The findings of GEBD can be applied to uniform sampling a video into a fixed number of frames based on generic boundaries to achieve high classification accuracy~\cite{Shou2021}. It can also provide a cue to select a set of frames for video summarization. Furthermore, knowledge of GEBD is needed for understanding in downstream tasks~\cite{wang2022gebPlus}. For example, boundary captioning means to generate description for the status change at the boundary, or boundary grounding that required to locate the boundary based on given description, or boundary caption-video retrieval aims to retrieve the video containing boundaries with provided description from video corpus.

Due to its taxonomy-free property, GEBD contains a variety of spatial and temporal characteristics that lead to difficulties in localizing these events as human perception.
Still in an early stage, only a few studies have been conducted to solve the GEBD problem. In previous studies~\cite{Shou2021,Tang_2022_CVPR}, the authors considered a set of consecutive frames as input, and its label corresponds to the label of the center frame. Although it reduced the computational burden for a single forward pass, but needed to repeat that process many times for the whole video and also lacked the relationships between frames in the longer view. Besides that, Li~\etal~\cite{Li_2022_CVPR} worked on the entire video but in the compressed domain, which caused deficiencies and reduced their performance.

In this study, we consider the entire video as an input that takes advantage of the complete semantic relationship between frames to localize event boundaries, and divide the video into meaningful units. First, based on the observation that the output from different layers of CNN-based architecture contains its characteristic to distinguish with the others, we collected features for each frame on multiple scales of spatial dimension, with a CNN backbone on its way to downsampling the frame. Second, 
motivated by the fact that two adjacent frames are almost similar in most of the cases in a video, we compute the similarity between adjacent frames by exploiting the correlation between frames in short and longer views together with different dilation rates in 1D temporal convolutions. In addition, our work also exploits the pyramid temporal relationship of the similarities to decode them into boundary scores.
In summary, the main contributions of this study are as follows:
\begin{itemize}
    \item We exploited the characteristic at different spatial dimensions to distinguish frames at multiple levels.
    \item Taking advantage of dilation rate in 1D convolution operation, we propose to use temporal pyramid similarity (TPS) module that builds upon this operation with dilation rates to construct the similarity between frames in multiple views to solve the GEBD problem.
    \item To enrich the information in higher views, a residual connection is used to integrate spatial and temporal information, and a 1D depthwise convolution is inserted at the beginning of these views to incorporate local information to them without increasing too much computational cost.
    \item Thanks to stacking multiple 1D convolution layers with different dilation rates to decode the similarity representation, our framework obtained improvements in performance compared to the state-of-the-art approach.
    
\end{itemize}
\section{Related Works}

\paragraph{Temporal modeling for spatial features} Although 3D convolution is designed as a native architecture to process a sequence of images~\cite{ji20123d,tran2015learning,Carreira_2017_CVPR}, but it takes a lot of computing resources. Leveraging 2D CNN to extract spatial features and then modeling temporal relationship with other architectures (e.g., 1D CNN, LSTM), is one approach to tackle that problem.
The authors in~\cite{Yu2019} and~\cite{Zheng2019} partitioned feature map to various levels and applied pooling operation on partitions, then concatenate results to achieve temporal pyramid features.
Wang~\etal~\cite{Wang2019} was inspired by DeepLab~\cite{chen2017deeplab,chen2018encoder}, they deployed multiple parallel temporal convolutions with different atrous rates to capture multiscale contextual features.
MS-TCN++~\cite{li2020ms} stacked multiple blocks of dual dilated temporal convolutions to capture temporal dependencies in a sequence.

\paragraph{Temporal activity detection in video} Early methods~\cite{shou2016temporal,dai2017temporal,chao2018rethinking,Lin2019} formulate the task to proposal generation and action classification. The generation of action proposals aims to find candidate segments, including a pair of starting and ending boundaries, in a long video that may contain actions. Shou~\etal~\cite{shou2016temporal} deployed 3D convolution operations to classify segments generated from sliding windows of varied length into two nodes (action or background). The authors in~\cite{dai2017temporal,chao2018rethinking,Lin2019} used 1D convolution operations to capture context information between features sampling on multiple scales. TURN~\cite{gao2017turn} divided a long video into video units and calculated CNN features for each unit. TURN employed an anchor unit to construct clip pyramid features and compute confidence scores. MGG~\cite{liu2019multi} deployed multiple perspectives of granularity, depending on visual characteristics with embedded position information. Yang~\etal~\cite{yang2022temporal} employed background constraints that take advantage of the rich information about the action and background to suppress low quality proposals.

\paragraph{Generic event boundary detection}
 Shou~\etal~\cite{Shou2021} introduced the first benchmark dataset to detect generic event boundaries, based on Kinetics-400 videos~\cite{Zisserman_k400}. They extended the boundary matching (BM) mechanism~\cite{Lin2019}, which considered each pair of starting and ending boundaries as a proposal and computed their confidence scores, to determine these boundaries. The authors in~\cite{Kang_2022_CVPR} exploited the pairwise self-similarity between video frames to construct a temporal self-similarity matrix (TSM) for the entire video to use as an intermediate stage and integrated with contrastive learning to produce boundary indices. Li~\etal~\cite{Li_2022_CVPR} encoded GOPs (group of pictures separated from video with modern codecs) based on their I-frames and P-frames to form unified representations and explore temporal dependence with the temporal contrastive module. DDM-Net~\cite{Tang_2022_CVPR} leveraged multi-level features of space and scale for calculating the difference at multiple scales and construct multi-level dense difference maps. These maps are then aggregated and fused with progressive attention.
 
\section{Method}
\begin{figure*}[htbp]
    \centering
    \includegraphics[scale=.85]{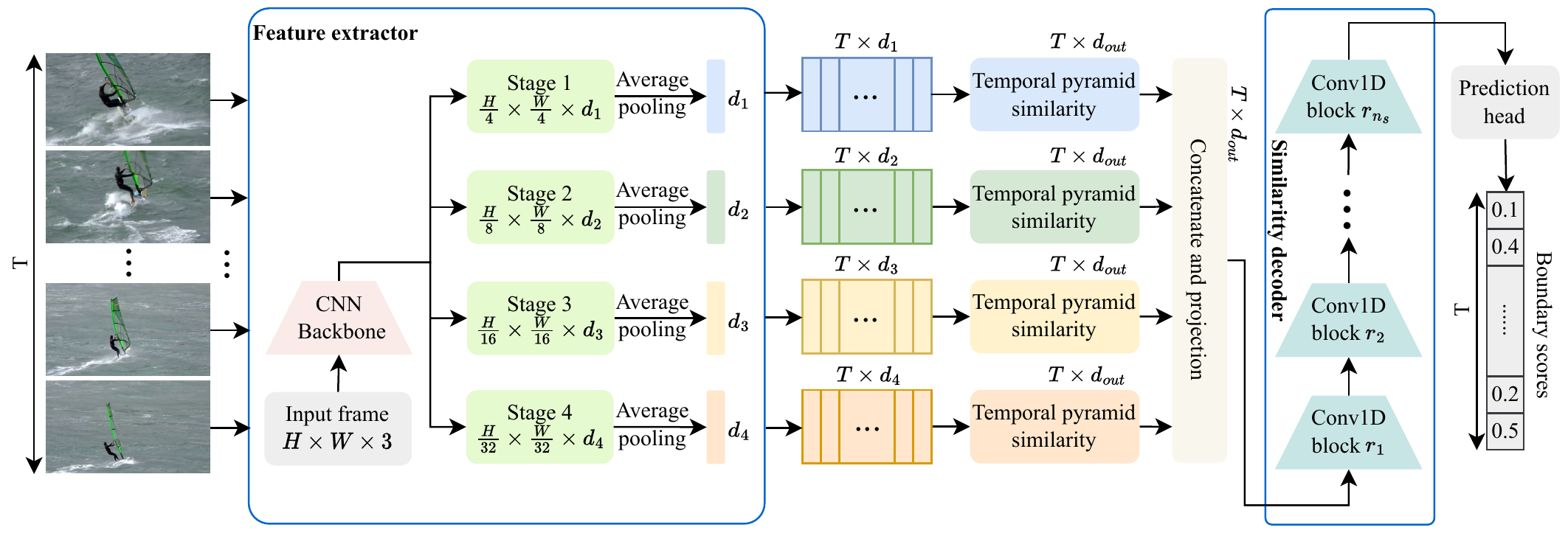} 
    \caption{An overview of the proposed system.}
    \label{fig:overview}
\end{figure*}

To estimate the boundary score for every frame in a video with $T$ frames, the proposed systems consider the similarity between each frame and their neighbors with the encoded features in a multi-view manner. These similarities are then decoded with 1D CNNs and post-processed with a Gaussian filter for smoothing and reducing noise on the estimated scores. An illustration of our system is shown in~\cref{fig:overview}.

\subsection{Spatial Pyramid Features}
Over the last decade, many deep learning architectures have been developed to tackle the image classification task, which can be used as a base for the other downstream tasks (e.g., activity recognition, video classification), with continuous improvement over time. Some of them are considered standard architectures when constructing the baseline model to solve the other problems. Following prior works~\cite{Shou2021,Kang_2022_CVPR,Tang_2022_CVPR,Li_2022_CVPR}, our network is built upon ResNet-50 architecture~\cite{he2016deep}, pre-trained on ImageNet, to construct spatial features for each frame in video. Given a frame resized to $224\times224$, our model leverages the output from the last convolution blocks in which the input is downsampled in a pyramid scheme by 4, 8, 16 and 32 times as they contain the most detail and vital information for each size of spatial dimensions. The average pooling is applied on the spatial dimensions of these outputs to obtain the 1D feature vector for each of them, as illustrated in \textit{Feature extractor} part of~\cref{fig:overview}. Based on that, this work takes advantage of 4 feature vectors of size $T\times d_{i}$ for a video with $T$ frames, where $d_{i}=2^{i+7}$ with $i\in\{1,2,3,4\}$ as produced by ResNet-50.

\subsection{Temporal Pyramid Similarity}
Based on the observation that two adjacent frames are almost similar in most cases when recording video, we perform 1D convolution operations with different dilation rates separately to achieve multiple views changing from short- to long-term correlation. Given a sequence $S$ and a filter $K$ of size $2m+1$ with $m\in\mathbb{N}$, convolution $G$ of these inputs with dilation rate $r$ is formulated as follows:
\begin{equation}
    G_{i}^{r} = \sum_{j=-m}^{m}K_{j}\cdot S_{i+r\cdot j}
\end{equation}
where $i+r\cdot j$ accounts for the dependence on the past and future of an output at time $i$. This work deploys $n=4$ blocks with kernel size of $3$ and dilation rates $r_{i}=2^{i-1}$ with $i\in\{1,2,...,n\}$, implies the system can observe up to $8$ frames in the temporal dimension to form the representations for obtaining the similarities between frames and their neighbors. To exploit the richer relationship between short-term and long-term correlation, for blocks with $r_{i}>1$, a 1D depthwise convolution~\cite{chollet2017xception} is added at the beginning of the block to form the adjacent relationship (local correlation), as presented in~\cref{tab:tps_conv1d} with $ks$ is kernel size and $r$ is dilation rate.
\begin{table}[htbp]
    \centering
    \caption{Details of each Conv1D block $r_{i}$ in TPS module.}
    \label{tab:tps_conv1d}
    \begin{tabular}{l c c c c} \toprule
         Layer & $ks$ & $r$ & Included & Output size \\\midrule
         input & $-$ & $-$ & \checkmark & $T\times d_{k}$ \\ 
         depthwise conv1d & 3 & 1 & $r_{i}>1$& $T\times d_{k}$ \\
         conv1d & $3$ & $r_{i}$ & \checkmark & $T\times d_{k}$ \\
         layer norm & $-$ & $-$ & \checkmark & $T\times d_{k}$ \\
         activation (gelu) & $-$ & $-$ & \checkmark & $T\times d_{k}$ \\
         \bottomrule
    \end{tabular}
\end{table}

\begin{figure}[htbp]
    \centering
    \includegraphics[scale=.85]{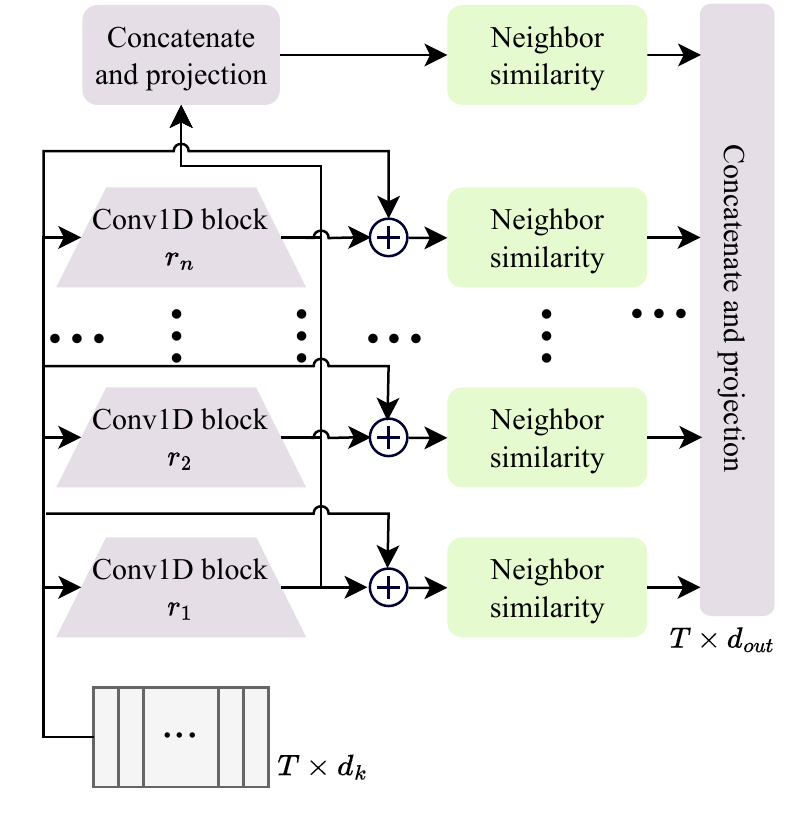}
    \caption{Temporal pyramid similarity module.}
    \label{fig:tpd_module}
\end{figure}

In addition, the output from blocks is also concatenated and projected to the lower dimension as the same as input, in which the most important properties are compressed to produce a comprehensive representation. In total, the system uses $n+1$ feature vectors with each size of $T\times d_{k}$, $k\in\{1,2,3,4\}$ corresponding to 4 stages in the feature extractor step, to calculate the similarity between neighbor frames with Euclidean distance, as done in~\cite{Tang_2022_CVPR}, for each stage, as shown in~\cref{fig:tpd_module}. To incorporate the difference in spatial dimension, we add a residual connection to connect each stage output with its input before measuring the similarity between frames. Formally, let $F_{k}^{i}$ is the output of the feature vector $I_k$ at stage $k$ with dilation rate $r_{i}$, we obtain $n$ similarity vectors by the following equation
\begin{align}
    R_{k}^{i} &= \frac{F_{k}^{i} + I_{k}}{\lVert F_{k}^{i} + I_{k} \rVert}, \\
    \mathcal{D}(q) &= \lVert R_{k, t}^{i} - R_{k, t+q}^{i} \rVert^{2}, \\
    D_{k}^{i} &= [\mathcal{D}(-l), \dots, \mathcal{D}(-1), \mathcal{D}(1), \dots, \mathcal{D}(l)],
\end{align}
with $\mathcal{D}(q)$ denotes the Euclidean distance between frame at time $t$ and $t+q$. In this work, we consider neighbors in 1 second for each side, in other words, $l$ equals the frames per second of video. And $R_{k}^{n+1}$, the comprehensive representation of $F_{k}^{i}$, is calculated as
\begin{align}
    P &= \mathcal{F}_p([F_{k}^{1}, F_{k}^{2}, \dots, F_{k}^{n}]), \\
    R_{k}^{n+1} &= \frac{P}{\lVert P \rVert},
\end{align}
where $\mathcal{F}_p$ is a 1D convolution with kernel size of $1$, and $P$ has the same dimension with $F_{k}^{i}$. These similarities, $R_{k}$, are concatenated and projected into higher dimension, 
\begin{equation}
    \mathcal{T}_{k} = \mathcal{F}_{t}([R_{k}^{1},R_{k}^{2},\dots,R_{k}^{n},R_{k}^{n+1}]),
\end{equation}
with $\mathcal{F}_t$ is 1D convolution with $d_{out}$ filters of kernel size of $1$. Consequently, this module formed a feature vector, $\mathcal{T}_{k}$, of size $T\times d_{out}$ for each stage output. At this point, we concatenate the feature vectors of 4 stages and project to a new dimension to obtain a final feature vector, $\mathcal{T}$, as
\begin{equation}
    \label{eq:tps_output}
    \mathcal{T} = \mathcal{F}([\mathcal{T}_{1}, \mathcal{T}_{2}, \mathcal{T}_{3}, \mathcal{T}_{4}]),
\end{equation}
in which, the projection $\mathcal{F}$ is a 1D convolution block with $d_{out}$ filters of kernel size $3$ with aim to  reduce the dimension of the concatenation vector and incorporate these information together.

\subsection{Similarity Decoder}
To decode the similarity information for further prediction, $n_{s}$ 1D convolution blocks with dilation rates $r_{i}$, $i\in\{1,2,...,n_{s}\}$, are stacked together to form a similarity decoder (SD) block. Due to dilation rate of $1$ is applied to the previous projection block, this module employs dilation rates greater than $1$ and does not include 1D depthwise convolutions. As 1D convolutions are stacked instead of applying separately as in the previous module, this decoder can observe more frames to compute the output at each timestamp. For simplicity, the same dilation rates as in the temporal pyramid similarity module are used. In detail, $n_{s}=n-1$ for this part and $r_{i}=2^{i}$. Formally, the output $\mathcal{S}$ of SD block is defined as
\begin{equation}
    \mathcal{S} = \left( \mathcal{F}^{r_{n_{s}}}\circ \mathcal{F}^{r_{n_{s}-1}} \circ \cdots \circ \mathcal{F}^{r_{1}} \right)(\mathcal{T}),
\end{equation}
with $\mathcal{T}$ from~\cref{eq:tps_output}, and $\mathcal{F}^{r_{i}}$ denotes the 1D convolution block with dilation $r_{i}$.

\subsection{Prediction Head}
Our network uses two layers of 1D convolutions, followed by a sigmoid function, to estimate the boundary score for each frame. These scores tend to be noisy and difficult to use only a fixed-threshold (e.g., 0.5) to get exact boundaries. To address these issues, we use a Gaussian kernel with standard deviation of $1$ and window size equal to the number of frames in 1 second, as shown in~\cref{fig:gf_prediction_viz}. Then each score is compared to neighbors in $0.5$ seconds and a minimum threshold of $0.1$, which works similar to max pooling, to consider as a maximum (boundary) or not.

\section{Experimental Results}

\begin{table*}[htbp]
	\centering
	\caption{Kinetics-GEBD validation score with ResNet50 as the backbone at Rel.Dis.\label{tab:kinetics-gebd-val}}
	\resizebox{\linewidth}{!}{%
	\begin{tabular}{l c c c c c c c c c c c} \toprule
		Method & F1@0.05 & F1@0.1 & F1@0.15 & F1@0.2 & F1@0.25 & F1@0.3 & F1@0.35 & F1@0.4 & F1@0.45 & F1@0.5 & Avg  \\ \midrule
		BMN~\cite{Lin2019}  & 0.186 & 0.204 & 0.213 & 0.220 & 0.226 & 0.230 & 0.233 & 0.237 & 0.239 & 0.241 & 0.223\\ 
		BMN-StartEnd~\cite{Lin2019} & 0.491 &  0.589 & 0.627 & 0.648 & 0.660 & 0.668 & 0.674 & 0.678 & 0.681 & 0.683 & 0.640 \\
		TCN~\cite{lea2016segmental}  & 0.588 &  0.657 & 0.679 & 0.691 & 0.698 & 0.703 & 0.706 & 0.708 & 0.710 & 0.712 & 0.685 \\
		PC~\cite{Shou2021}  & 0.625 &  0.758 & 0.804 & 0.829 & 0.844 & 0.853 & 0.859 & 0.864 & 0.867 & 0.870 & 0.817 \\
		SBoCo~\cite{Kang_2022_CVPR} & 0.732 & 0.827 & 0.853 & 0.877 & 0.882 & 0.891 & 0.894 & \textbf{0.899} & 0.899 & \textbf{0.907} & 0.866 \\
		DDM-Net~\cite{Tang_2022_CVPR} & 0.764 & \textbf{0.843} & \textbf{0.866} & \textbf{0.880} & \textbf{0.887} & \textbf{0.892} & \textbf{0.895} & 0.898 & \textbf{0.900} & 0.902 & \textbf{0.873} \\
		CVRL~\cite{Li_2022_CVPR} & 0.743 & 0.830 & 0.857 & 0.872 & 0.880 & 0.886 & 0.890 & 0.893 & 0.896 & 0.898 & 0.865 \\ \midrule 
		Ours & \textbf{0.770} & 0.842 & 0.863 & 0.874 & 0.880 & 0.884 & 0.887 & 0.890 & 0.891 & 0.893 & 0.867 \\
		\bottomrule
	\end{tabular}
	}
\end{table*}

\begin{table*}[htbp]
	\centering
	\caption{TAPOS validation score compared to previous works.\label{tab:tapos-val}}
	\resizebox{\linewidth}{!}{%
	\begin{tabular}{l c c c c c c c c c c c} \toprule
		Method & F1@0.05 & F1@0.1 & F1@0.15 & F1@0.2 & F1@0.25 & F1@0.3 & F1@0.35 & F1@0.4 & F1@0.45 & F1@0.5 & Avg  \\ \midrule
		ISBA~\cite{ding2018weakly} & 0.106 & 0.170 & 0.227 & 0.265 & 0.298 & 0.326 & 0.348 & 0.369 & 0.382 & 0.396 & 0.302 \\
		TCN~\cite{lea2016segmental}  & 0.237 & 0.312 & 0.331 & 0.339 & 0.342 & 0.344 & 0.347 & 0.348 & 0.348 & 0.348 & 0.330 \\
		CTM~\cite{huang2016connectionist} & 0.244 & 0.312 & 0.336 & 0.351 & 0.361 & 0.369 & 0.374 & 0.381 & 0.383 & 0.385 & 0.350 \\
		TransParser~\cite{shao2020intra} & 0.289 & 0.381 & 0.435 & 0.475 & 0.500 & 0.514 & 0.527 & 0.534 & 0.540 & 0.545 & 0.474 \\
		PC~\cite{Shou2021}  & 0.522 & 0.595 & 0.628 & 0.646 & 0.659 & 0.665 & 0.671 & 0.676 & 0.679 & 0.683 & 0.642 \\
		DDM-Net~\cite{Tang_2022_CVPR} &  0.604 & 0.681 & 0.715 & 0.735 & 0.747 & 0.753 & 0.757 & 0.760 & 0.763 & 0.767 & 0.728 \\
		\midrule 
		Ours &  \textbf{0.616}  &  \textbf{0.701} &  \textbf{0.740} & \textbf{0.760} & \textbf{0.772} & \textbf{0.780} & \textbf{0.786}  & \textbf{0.790} & \textbf{0.794}  & \textbf{0.796} & \textbf{0.754} \\
		\bottomrule
	\end{tabular}
	}
\end{table*}

\subsection{Dataset and Evaluation}
\paragraph{Kinetics-GEBD}
Our approach is evaluated primarily on Kinetics-GEBD, a benchmark dataset for locating the boundaries of generic events in video, created by Shou~\etal~\cite{Shou2021}. It consists of around \num{60000} 10-second videos, with the training, validation, and testing partition ratio approximately 1:1:1. The training and testing videos are randomly selected from Kinetics-400~\cite{Zisserman_k400} Train split, while the validation set used all videos in Kinetics-400 Val split. Each video is annotated by 5 annotators with average 4.77 boundaries. Due to the annotation of test set not available publicly, our network is trained on training set and evaluated on validation set.

\paragraph{TAPOS} In addition, we conduct experiment on TAPOS dataset~\cite{shao2020intra} containing Olympics sport videos across 21 actions. Follow~\cite{Shou2021}, we perform boundaries localization between sub-actions in each action instance, as annotated in~\cite{shao2020intra}. The average duration of instances is 9.4 seconds, and maximum around 5 minutes.
TAPOS contains \num{13094} action instances for training and \num{1790} instances for validation. Due to the variety of duration over instances, we split instances into 10-second clips with overlap of 5 seconds to perform training and evaluation. During evaluation, the boundary scores from clips are merged with summation to form a unique score for each timestamp in the whole action instance to get the final boundaries.
Similar to Kinetics-GEBD, our network is trained on the training set and evaluated on the validation set.

\paragraph{Evaluation}
Following~\cite{Shou2021}, F1 score with Relative Distance (Rel.Dis.) measurement is used to evaluate the system performance. Rel.Dis. determines whether a detection is correct or incorrect by calculating the error of detection, $E_{d}$, as
\begin{equation}
    E_{d} = \frac{\lvert\mathrm{Detected~point} - \mathrm{Ground~truth~point}\rvert}{\mathrm{Length~of~video}},
\end{equation}
and comparing with a threshold, $\tau$. If $E_{d}\leq\tau$, the detection is correct, otherwise incorrect. The reported results are with thresholds $\tau$ from $0.05$ to $0.5$ with a gap of $0.05$.

\subsection{Implementation}
Because of the variety of frame rate of each video, we sample frames at 25 fps (frame per second), as done in~\cite{Carreira_2017_CVPR}. For this task, video at 25 fps may contain redundant information. We then sample video to 5 fps that led to 50 frames per video or $T=50$, which is suitable to run the inference on a usual computing resource. The network is built with TensorFlow 2.9 and trained end-to-end with ResNet-50 backbone and Adam optimizer at batch size of $8$ in 10 epochs. We linearly increased the learning rate from 0 to \num{4e-4} in 2 epochs, and then used a cosine decay schedule to reduce the learning rate to \num{4e-6} in 8 epochs. The model from the last epoch is used to generate the prediction for evaluation. All experiments are conducted on a single NVIDIA RTX 3090 GPU equipped machine.

\subsection{Results}
\Cref{tab:kinetics-gebd-val} shows the comparison of our approach with the previous method on the Kinetics-GEBD validation set with the same spatial backbone (ResNet-50). Compared to PC~\cite{Shou2021}, the GEBD benchmark baseline, our system achieves a huge improvement of almost $14.5$ percent, demonstrating the effectiveness of the proposed system. DDM-Net~\cite{Tang_2022_CVPR} took consecutive frames with a gap between them that created a longer view in their system. Consequently, they also surpass the baseline, but still has a small gap, $0.6$ percent, with our method at strict threshold $\tau=0.05$ as multiple longer views are exploited in our method. We also achieved a remarkable improvement compared to CVRL~\cite{Li_2022_CVPR} and SBoCo~\cite{Kang_2022_CVPR} with around $3$ percent at $\tau=0.05$.

The results of the comparison between our method and previous methods on TAPOS are summarized in~\cref{tab:tapos-val}. Since our approach is able to learn different context in multiple view, we surpass PC~\cite{Shou2021} and DDM-Net~\cite{Tang_2022_CVPR} as TAPOS contains instances with duration up to 5 minutes. We obtain an increasing of around $8.2\%$ and $1.2\%$ on F1@0.05 compared to PC and DDM-Net, respectively. These results prove the effectiveness of our method on short videos, around 10 seconds, and long videos with up to 5 minutes.


\subsection{Ablation study}
To evaluate the effectiveness of our approach, several ablations on the network are conducted with the same training mechanism.

\paragraph{Effects of stages in the feature extractor} 
We experiment our system with different number of feature maps to analyze how it effects the overall performance. As high-level feature maps (high stages) contain more specific information compared to low-level feature maps (low stages), we add/remove feature maps from high to low in an orderly way as in~\cref{tab:eff_stages}. The system has increased steadily in terms of F1@0.05 with a difference of up to $1.22$ percent between the single stage and the combination of 4 stages. This can be explained because each feature map level contains its own knowledge that makes the different neighboring frames, which makes their combination useful for the whole system.

\begin{table}[htbp]
    \centering
    \caption{Effects of the combination of stages in feature extractor with results on Kinetics-GEBD.}
    \label{tab:eff_stages}
    \begin{tabular}{l c c c} \toprule
        Stage & Prec@0.05 & Rec@0.05 & F1@0.05 \\ \midrule
        Stage 4 & 0.6965 & 0.8330 & 0.7587 \\
        Stage 3, 4 & 0.6967 & 0.8462 & 0.7642 \\ 
        Stage 2, 3, 4 & 0.7026 & 0.8460 & 0.7677 \\
        Stage 1, 2, 3, 4 & \textbf{0.7076} & 0.8450 & \textbf{0.7702} \\ 
        Stage 1, 2, 3 & 0.7015 & 0.8491 & 0.7683 \\ 
        Stage 1, 2 & 0.6956 & \textbf{0.8514} & 0.7657 \\
        Stage 1 & 0.6924 & 0.8373 & 0.7580 \\
        \bottomrule
    \end{tabular}
\end{table}

\paragraph{Effects of dilation rates} \Cref{tab:eff_dlr} shows the results when adding dilation rates of $2^{i-1}$ to compute the similarity in TPS module. In general, F1 scores increase consistently around $0.1$ percent for each new dilation rate and reach a total gain of $0.35\%$ based on the improvement $0.5\%$ of precision. Although these increases are not remarkable, they still show a potential further extension of how to incorporate them in an efficient manner and achieve significant improvements.

\begin{table}[htbp]
    \centering
    \caption{Effects of dilation rates in the TPS module with results on Kinetics-GEBD.}
    \label{tab:eff_dlr}
    \resizebox{\linewidth}{!}{%
    \begin{tabular}{l c c c} \toprule
         Modification & Prec@0.05 & Rec@0.05 & F1@0.05 \\ \midrule
         $n=1, r_{i}=2^{i-1}$ & 0.7013 & 0.8455 & 0.7667 \\
         $n=2, r_{i}=2^{i-1}$ & 0.7045 & 0.8438 & 0.7679 \\
         $n=3, r_{i}=2^{i-1}$ & 0.7037 & \textbf{0.8472} & 0.7688 \\
         $n=4, r_{i}=2^{i-1}$ & \textbf{0.7076} & 0.8450 & \textbf{0.7702} \\
         \bottomrule
    \end{tabular}
    }
\end{table}

\paragraph{Effects of local correlation} In this study, we incorporate residual connection (Res.Con.) and 1D depthwise convolution to enrich long-term correlation with local contextual information in the TPS module.~\Cref{tab:eff_local_corr} shows their effects on overall system performance. According to that, Res.Con. achieves an increase of $1.2\%$,  while the other does not have many effects. Consequently, it encourages refinement of features with local context in a whole view with Res.Con. to improve the generalization instead of on individual channels as to what depthwise convolution is doing.

\begin{table}[htbp]
    \centering
    \caption{Effects of local correlation in TPS module with results on Kinetics-GEBD, in which Res.Con. indicates residual connection.}\label{tab:eff_local_corr}
    \resizebox{\linewidth}{!}{%
    \begin{tabular}{M{.15\linewidth} M{.12\linewidth} c c c} \toprule
        Depthwise conv1d & Res.Con. & Prec@0.05 & Rec@0.05 & F1@0.05 \\ \midrule
        & & 0.6943 & 0.8361 & 0.7586 \\
         & \checkmark & 0.7063 & \textbf{0.8470} & \textbf{0.7703} \\
        \checkmark & & 0.6933 & 0.8373 & 0.7585 \\
        \checkmark & \checkmark & \textbf{0.7076} & 0.8450 & 0.7702 \\ \bottomrule
    \end{tabular}
    }
\end{table}
\paragraph{Effects of similarity decoder block}
We conduct experiments with a fixed dilation rate in SD block along with the removal of SD block to compare their results, as shown in~\cref{tab:eff_sd}. Generally, SD block helps improve performance at least $0.8\%$ with no dilation and at most $1.2\%$ with pyramid dilation rates. The dilation rates can improve the performance, but large dilation can also cause the drop. In case of the pyramid dilation rates, it creates balances by increasing the receptive field step by step, which consistently boosts the performance.

\begin{table}[htbp]
    \centering
    \caption{Effects of similarity decoder block with results on Kinetics-GEBD. In these experiments, $n_{s}$ is set to $3$.}
    \label{tab:eff_sd}
    \resizebox{\linewidth}{!}{%
    \begin{tabular}{l c c c} \toprule
         Modification & Prec@0.05 & Rec@0.05 & F1@0.05 \\ \midrule
         Excluded SD block & 0.6998 & 0.8280 & 0.7585 \\
         SD block $r_{i}=1$ & 0.7059 & 0.8395 & 0.7669 \\
         SD block $r_{i}=2$ & 0.7036 & \textbf{0.8462} & 0.7684 \\
         SD block $r_{i}=4$ & 0.7059 & 0.8443 & 0.7689 \\
         SD block $r_{i}=8$ & 0.7060 & 0.8333 & 0.7644 \\
         SD block $r_{i}=2^{i}$ & \textbf{0.7076} & 0.8450 & \textbf{0.7702} \\
         \bottomrule
    \end{tabular}
    }
\end{table}

\begin{table}[htbp]
    \centering
    \caption{Effects of Gaussian smoothing in prediction head with results on Kinetics-GEBD.}
    \label{tab:eff_gaussian}
    \resizebox{\linewidth}{!}{%
    \begin{tabular}{c c c c c } \toprule
         \multicolumn{2}{c}{Gaussian smoothing} & \multirow{2}{*}{Prec@0.05} & \multirow{2}{*}{Rec@0.05} & \multirow{2}{*}{F1@0.05} \\ \cmidrule{1-2}
         Training & Inference & & & \\ \midrule
         &  & 0.6717 & \textbf{0.8631} & 0.7555 \\
         & \checkmark  & \textbf{0.7426} & 0.8038 & \textbf{0.7720} \\ 
         \checkmark &  & 0.6206 & 0.9103 & 0.7381 \\
         \checkmark & \checkmark  & 0.7076 & 0.8450 & 0.7702 \\ \bottomrule
    \end{tabular}
    }
\end{table}

\begin{figure*}[htbp]
    \centering
    \includegraphics[width=.9\linewidth]{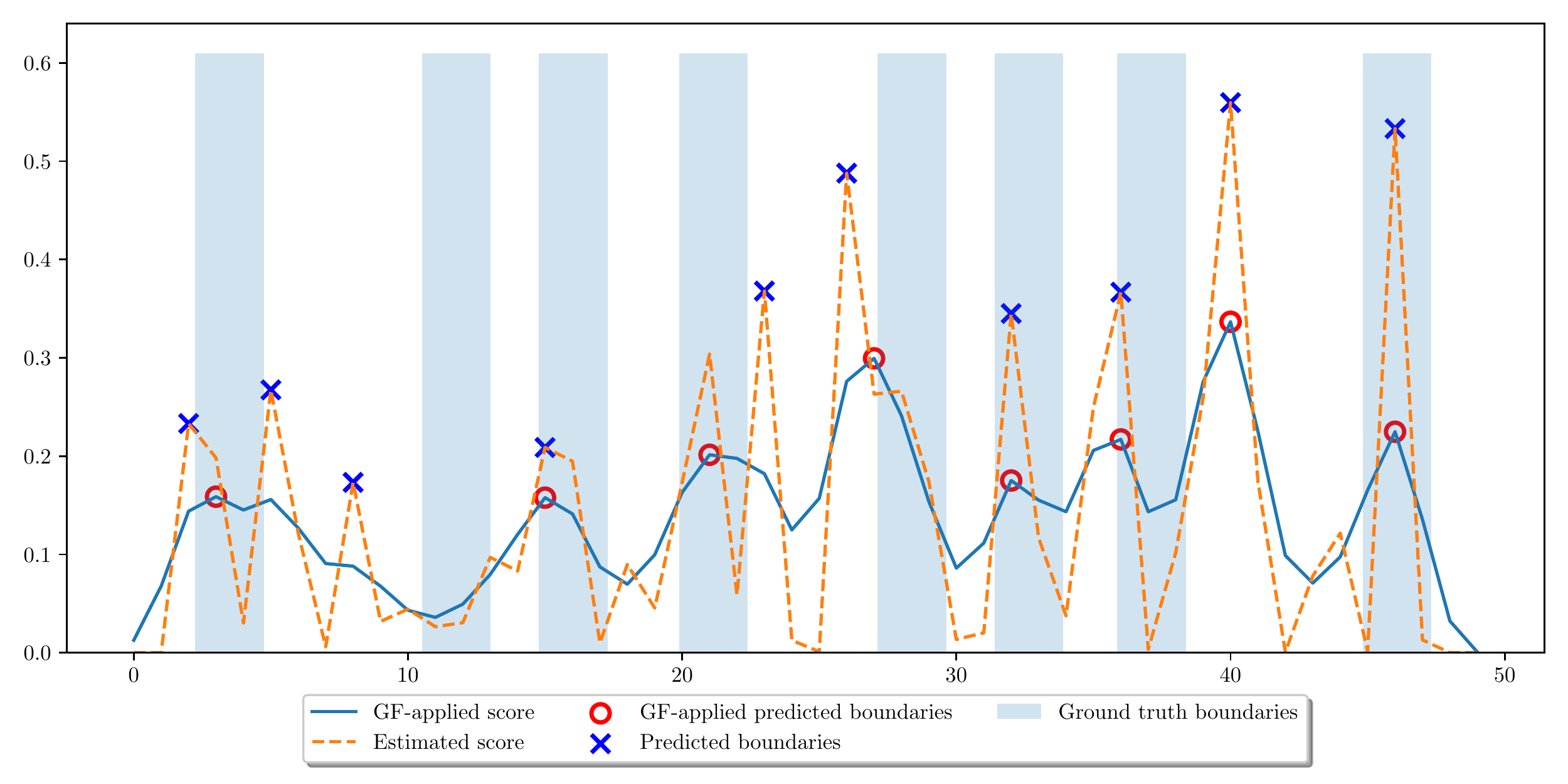} 
    \caption{A visualization of detected boundaries with or without GF smoothing.}
    \label{fig:gf_prediction_viz}
\end{figure*}
\paragraph{Effects of Gaussian smoothing in prediction head}
In this work, the Gaussian filter (GF) is attached and runs on the GPU together with the other parts of the network during training and inference. If GF includes in training, the learning process will optimize the smoothing scores instead of the raw ones. To see its impacts on the overall performance, we do include and exclude it during training and inference, respectively. As we can see in~\cref{tab:eff_gaussian}, the recall is very high compared to precision when GF is excluded for inference, which makes more noise in scores with many local maximums 
and produces many boundaries. On the other hand, the network focus on the smoothing score and cause a huge loss, nearly $3.21\%$ that twice compared to the other ($1.65\%$).~\Cref{fig:gf_prediction_viz} shows an example result in which GF reduced 4 false positive detection with Rel.Dis. threshold of $0.05$.

\section{Conclusion}
In this work, we proposed an approach based on temporal pyramid similarity to detect the boundary of generic events in video. Our work exploited the correlation between frames in different temporal and spatial scales in a pyramid scheme to compute the distance between adjacent frames, parallelly. Together, we applied this operation on multiple spatial scales to combine the unique characteristics on each scale to improve the system performance. Thanks to 1D operation for temporal modeling, we explored the temporal dimension to decode the similarity information and achieve a considerable improvement for the whole system. Our method outperforms the previous state-of-the-art methods on the Kinetics-GEBD and TAPOS benchmarks at a strict relative distance. More research is needed to expand our method so that it can work with untrimmed videos, as we only evaluate our approach for trimmed video with action.

{\small
\bibliographystyle{ieee_fullname}
\bibliography{egbib}
}

\end{document}